\pdfoutput=1
\documentclass[letterpaper]{article} 
\usepackage{aaai19}  
\usepackage{times}  
\usepackage{helvet}  
\usepackage{courier}  
\usepackage{url}  
\usepackage{graphicx}  
\usepackage{smartdiagram}  
\usepackage{tikz}
\usetikzlibrary{matrix}
\usepackage{stmaryrd}
\usepackage{braket}
\usepackage{amssymb}
\usepackage{amsmath}
\usepackage{mathrsfs}
\usepackage{mathbbol}
\usepackage{array}

\def\defemb#1#2{\expandafter\def\csname #1\endcsname
	{\relax\ifmmode #2\else\hbox{$#2$}\fi}}

\defemb{cD}{{\cal D}}
\defemb{cP}{{\cal P}}
\defemb{cV}{{\cal V}}

\newcommand{\state}{s}

\newcommand{\action}{o}

\usepackage{times}

\begin{document}
\title{Economics of Human-AI Ecosystem: Value Bias and Lost Utility in Multi-Dimensional Gaps}
\author{Daniel Muller\\
The Faculty of Industrial Engineering and Management\\
Technion - Israel Institute of Technology, Haifa, Israel\\
mullerdm@gmail.com\\
}
\maketitle
\begin{abstract}
In recent years, artificial intelligence (AI) decision-making and autonomous systems became an integrated part of the economy, industry, and society. The evolving economy of the human-AI ecosystem raising concerns regarding the risks and values inherited in AI systems. This paper investigates the dynamics of creation and exchange of values and points out gaps in perception of cost-value, knowledge, space and time dimensions. It shows aspects of value bias in human perception of achievements and costs that encoded in AI systems. It also proposes rethinking hard goals definitions and cost-optimal problem-solving principles in the lens of effectiveness and efficiency in the development of trusted machines. The paper suggests a value-driven with cost awareness strategy and principles for problem-solving and planning of effective research progress to address real-world problems that involve diverse forms of achievements, investments, and survival scenarios.

\end{abstract}
\section{Introduction}

Problem-solving and planning are decision-making processes that consist of ordered decision choices and decision actions~\cite{kriger1992organizational,marwala2015causality}. Instantaneous decision-choices are the atomic units that compose a decision-action. Decision-theoretic planning is an approach~\cite{boutilier1999decision} for solving sequential decision problems that result with a plan or a policy. Under the assumption of a deterministic environment, a course of actions is a plan that guarantees to reach a specific goal. A more flexible framework for a solution plan or policies is a course of action with an expected high utility that fits uncertainty scenarios and a more complex target structures and reward dynamics represented in preferences value or utility functions. An optimized decision results in a global maximization of utility \cite{degroot2005optimal,berger2013statistical}. Decision actions can be rational or irrational. Followed by~\cite{marwala2015causality} we define the decision action as {\bf rational} if it results in a global optimum, based on logical principles and derived from complete relevant information. A decision that based on irrelevant or incomplete information is irrational, and cannot guarantee guidance for a global optimum. Rational decision-making process comprises rational decision actions, and the entire process optimized in time and results in a global utility optimum~\cite{grune2012paradoxes,marwala2014artificial,marwala2015causality}

If so, rational decisions made only in a deterministic, perfect world with complete information. However, real-world scenarios are more complicated. It is not always possible to analyze all available and relevant information to evaluate all options required for making a responsible decision. For instance, planning domains with uncertainty, by default solved with an irrational decision-making process due to limited relevant information. Consider the deterministic oversubscription planning (OSP)~\cite{smith:icaps04,van2004effective,do2004partial,van2004over,nigenda2005planning,benton2006solving,aghighi2014oversubscription,DoBBK:ijcai07,mirkis:domshlak:jair15,muller:Karpas:icaps18}, dealing with domains in which there is over-subscription of possible achievements to limited resources. This problem is computationally challenging, scaling to real-world complexity with multi-valued arbitrary utility functions over achievements, numerical utility values are challenging even in small domains, due to limited processing capability for inference and utilization of relevant information. ~\citeauthor{aghighi2014oversubscription}~(\citeyear{aghighi2014oversubscription}) provided a detailed complexity analysis on the OSP problem. Despite the theoretically deterministic definition of the problem, limited computational capabilities do not allow us to enjoy determinism. This concept termed {\bf bounded rationality}~\cite{simon1957models,simon1990mechanism,simon1991bounded} and explains irrational decision making based on incomplete data analysis.~\cite{marwala2013flexibly} Extended this term to {\bf flexible-bounded rationality}, based on the concept that The advance of AI and machine data analysis makes the bounds of rationality flexible. The imperfection of information can be partially corrected by using advanced information analysis methods.

An important observation of~\cite{marwala2013flexibly} is that rationality is not dividable, even one irrational action makes the entire process irrational, which makes guarantees for optimality to irrational. A process of applying a sequence of depended actions cannot be partly rational and partly irrational. An imperfection of relevant information, leads to an entire decision process irrationality with a sub-optimal end, in time or utility. Scaling up to over-subscription planning with a multi-valued general utility function, we have to be concerned with the challenge of an exponential blow-up even in small tasks. We base our approach on~\cite{marwala2013flexibly} observation that the abundance of information with limited processing capability results in the incompleteness of effective information for a rational decision.

\begin{figure}[htb]
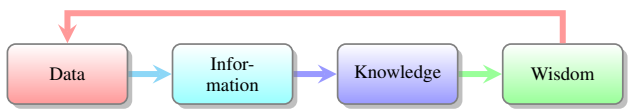

 \centering
 \resizebox{\columnwidth}{!}{%
  \smartdiagram[flow diagram:horizontal]{Data, Information, Knowledge, Wisdom}
 }
\caption{\label{fig:DIKW} DIKW hierarchy.}
\end{figure}

The relation of information and knowledge modeled in the in DIKW hierarchy (data - information - knowledge - wisdom) as illustrated in Figure~\ref{fig:DIKW}. Some researches agree that the origins of DIKW are in T.S. \citeauthor{eliot1934rock}'s poem The Rock (\citeyear{eliot1934rock}). Years later, the DIKW hierarchy got attention in the literature of research~\cite{zeleny1987management,cooley1980architect,ackoff1989data}. An in-deep survey by~\citeauthor{Jennifer2007Rowley} (\citeyear{Jennifer2007Rowley}) shows several observations and extensions to the basic model. \citeauthor{ackoff1989data} (\citeyear{ackoff1989data}) made an example in practice to the theory by providing a refined, one-page paper explanation to the relations between the DIKW hierarchy component and analysis of the effectiveness and efficiency aspects in a process to obtain knowledge and wisdom from data. The key concepts are as follows.

\begin{itemize}
\item \textbf{Data} are symbols that represent the properties of objects and events.
\item \textbf{ Information} is processed data directed at increasing its usefulness (utility), compactly described data.
\item \textbf{Knowledge} is the instructions on {\em how to} use information.
\begin{itemize}
\item Understanding is an explanation {\em why}.
\item Intelligence is the ability to increase efficiency, not effectiveness. Information, knowledge, and understanding constitute Intelligence.
\end{itemize}
\item \textbf{Wisdom} is the ability to increase effectiveness, to add value, which requires {\em judgment function}. Effectiveness is an evaluated efficiency.
\end{itemize}

In an imperfect world, we should be aware of the extent to which we allow missing information and its relevance for the decision process and the process of {\em turning knowledge into action}.
Judgment function which evaluates effectiveness purely leads to making unconscious mistakes or crimes by {\em turning information into action}.

In the literature of organizational management KHiA ``know-how-in-action" \cite{knowingCreatesValue2011,strati2007sensible,cunliffe2008orientations,empson2001fear} is a model of {\em turning knowledge into action}. Effective action depends on an effective judgment function must be based on knowledge. The effectiveness of actions varies with the context which depends on the specific people, the place and the time in which applied. Knowledge as a static resource, which creates value within a context and through actions, at a place in time \cite{knowingCreatesValue2011,spender1996organizational}. In what follows we define rationality as the \textit{ judgment function} of evaluating efficiency.

\section{Values Utilities and Trust}
Value is a state-related term quantifies some existence like goods, aggregated beliefs of society like money. Ethical values are the aggregated conception of an interrelated society regarding principles or norms. Future value is the potential of the mentioned above, an expected wealth. Values are not obsolete for few reasons; First society constructed from individuals, aggregated value of individuals is relatively fair since each weighted differently in aggregation. For example, the value of money and the derived pricing of goods is a function of the amount of its existence and availability. Since individuals posse different amount of money, they represented in the evaluation differently.

\begin{table*}[ht]
\centering
\resizebox{1.6\columnwidth}{!}{%
\setlength{\tabcolsep}{.5em}
    \begin{tabular}{|p{2.5cm}|p{5cm}|p{2.5cm}|p{5cm}|}
    \hline
    \multicolumn{2}{|c|}{\bf Dimensions of Value}   & \multicolumn{2}{c|}{\bf Dimensions of Costs}  \\ 
    \hline
    \hline
    {\bf Ethical Value} & principles and norms of a society & {\bf Ethical Costs }& reputation, trade of values, war, betrays, crime\\ \hline
    {\bf Financial Value} & currency, coins & {\bf Financial cost} & currency, coins\\ \hline
    {\bf Knowledge} & KHiA ``know-how-in-action", creation, productivity & {\bf Knowledge potential} & unrealized potential, obtained and shared (e.g. social good), obtained but not enrolled to value (e.g. stolen, lost)\\ \hline
    {\bf Labor value} & potential of value creation (not only financial, can be enrolled to any value) & {\bf Labor cost} & usually evaluated with money, but includes many human related issues like motivations and effort, time and more\\ \hline
    {\bf Emotional Value} & can be enrolled in to motivation for example and through that to creation of any value & {\bf Emotional cost} & implies decrease in motivation, low productivity\\ \hline
    {\bf Symbolic Value} & articles, desire, places, traditions, associated with memories and beliefs (related to past and future) & {\bf Symbolic cost} & losing hope, unstable connection to society, individualism\\ \hline
    {\bf The value of time} & can be invested as a cost of a new productive process acceleration of productive process, travel, surfing, opportunity to have benefit of a product, creativity, social bonding & {\bf The cost of time} & all what could happen but did not, unmaintained values, can be enrolled into decrease of each of the above mentioned values\\ \hline
    {\bf ...} &  & {\bf ...} & \\ \hline    
    \end{tabular}
   }
   \caption{\label{table:dimensions}
   Dimensions of costs and values}
\end{table*}

While the amount of money or price is obsolete, each evaluates it differently, from its perspective. Theory of labor states that people subjective to value relative to the troubles they have to suffer to achieve it. The literature addresses the concept of subjective evaluation with the utility term. In this paper, we express in the utility the usefulness of an action or process to create value. Useful action or process will make a net positive impact on total value and not useful will make a negative impact on the value. We distinguish between two types of utility.
\begin{enumerate}
\item Individual, subjective value of a productive process.
\item Social, aggregated value of the productive process.
\end{enumerate}
The first type of utility will be called merely utility to fit the traditional terms that explain this phenomenon. The literature treats mainly to the outcome of an action as the utility. The term marginal utility covered the change and defined as an added benefit with the increase of one unit of product. The definition of marginal utility is good enough to evaluate the local progress of a process, on the level of a single action and to observe the usefulness.

While this is enough to model some economic aspects, it does not model the real-world complexity of the change of value dynamics. Real economy involves more complex scenarios that cannot be measured marginally local. For example, an investment can take many forms and strategies and involve multidimensional value dynamics; time, place, emotional effort, investigation effort and more. The nature of investments is that they made with a future perspective, i.e., make a significant effort under limited resources for a greater good, and it should be planned responsibly. Due to time and uncertainty dynamics of investments, a trusted system and society becomes a significant issue. Without a system trust long-term projects cannot be done, and with short and efficient self-concentrated project it is only a matter of time that a system will come apart. The {\em human-AI ecosystem} that constitutes of individuals and groups of individuals and there slaved machines. Individuals that take individual, local in place and time perspective, will act cost-optimally, neglecting long-term implications. Cost optimality is dangerous when performed by people or bounded rational machines since optimality in cost is frequently achieved at the expense of values, both of the cost-optimal agent and other agents with which it interacts. If we can not promise complete rationality in our machines, the cost-optimal strategy must be re-examined for the sake of values that out of the boundaries of the perception of irrational machines.

Here we concentrate on two kinds of economic errors that are critical to any system that plans for trust. From an economic perspective, an autonomous agent or human agents are untrustable if they apply the following value-errors.
\begin{itemize}
\item An error in perception and awareness of self-actions contradicting the principles of the environment in which he acts with the impact the global system and its individual’s values.
\item The agent is trustable if he will not commit crimes of values, i.e., exploiting system and value gaps for arbitrage.
\end{itemize}

In what follows we will show several dimensions of value and analyze inter-dimensional gaps or traps.

\subsection{Hidden Dimensions of Costs and Values}

Recent studies showed interest in a non-monetary evaluation of goods and activities~\cite{poladian2003time,garrison2016value} presenting a more complex and realistic scheme of values. The dimensions of value and cost are defined by a society or organization and cannot be measured in the traditional monetary terms. The non-monetary dimensions of value are separated and measured on different scales. The dynamic between dimensions creates value from values, for example, the value of knowledge can be created by paying the cost in all the values as mentioned in Table\ref{table:dimensions}, which are enrolled into knowledge, unlike the traditional financial value and cost which paid and gained in the same monetary terms. When a discussion takes place on ethics and fairness, the dimensions must be separated and cannot be treated with different terms or subjectively as the utility term which is discussed in the next section.

\subsection{The Diamond-Water Paradox}
The utility is a subjective evaluation from an individual perspective of the wealth that he gains from a specific value. The paradox of value (and utility) as presented by \cite{smith1776inquiry};
\begin{quotation}
{\em The things which have the greatest value in use have frequently little or no value in exchange; and, on the contrary, those who have the greatest value in exchange have frequently little or no value in use. Nothing is more useful than water: but it will purchase scarce anything; scarce anything can be had in exchange for it. A diamond, on the contrary, has scarce any value in use; but a very great quantity of other goods may frequently be had in exchange for it. }
\end{quotation}
The labor theory \cite{farber2002economic} of value solve this paradox by explaining the cost of everything that one wants with the relative difficulty to fulfill the want, and the difficulties can be different as Figure~\ref{table:UtilityTypes} shows. Note, autonomous, AI and information systems produce some utility types that listed in Table~\ref{table:UtilityTypes}. For example, the development of the internet combined with the development of the mobile phone created through the form utility, utilities of different types. Mobile phone with Internet connection solves many issues of things we want in place and time. At the same time, the same phone is the reason that working days continue after working hours, this is the duality of availability.
\begin{table*}[htb]
\centering
\resizebox{1.6\columnwidth}{!}{%
\setlength{\tabcolsep}{.5em}
    \begin{tabular}[c]{|m{1.8cm}|m{4cm}|m{4cm}|m{4cm}|}
    \hline
    \multicolumn{2}{|c|}{\centering \bf Creating Utilities} & \multicolumn{2}{|c|}{\centering \bf Examples}\\
    \hline
    \hline
    \multicolumn{1}{|c|}{\centering \bf Types} & \multicolumn{1}{|c|}{\centering \bf Definition} & \multicolumn{1}{|c|}{\centering \bf  $\uparrow$} &  \multicolumn{1}{|c|}{\centering \bf  $\downarrow$ }\\
    \hline    
            {\bf Form } &
            created by design / reshape / reformulate / assemble of goods or services &
            software upgrade, marketing reshape information  &  error in upgraded software, waste
            \\ 
    \hline
            {\bf Place } & 
            created by making goods and services easier to obtain {\bf where} they needed (deliverable) & 
            markets, stores, selling water at a mountain peak   &  Missing stock a store, bank that can not give a loan
            \\ 
    \hline
            {\bf Time } &  
            created by making goods and service easier to obtain {\bf when} they needed & 
            fast-food, opening hours of stores  & Absence of emergency medical services
            \\ 
    \hline
            {\bf Task, Knowledge, Service } & 
            created by providing professional and personal service, assistance, guidance, etc. & 
            bankers, layers, academic advisor, ministers, reviews, doctors, etc.  & supplying bad service, false and non-promoting tips, bureaucracy
            \\ 
    \hline
            {\bf Possession } & created by transfer of ownership 
            & car
            & dept 
            \\ 
    \hline
            {\bf Information } & communication
            & 
            marketing, teaching, answering questions, navigation directions etc.  & false information
            \\ 
    \hline
    \end{tabular}
}
   \caption{\label{table:UtilityTypes}
   Types of utilities and examples for creating positive and negative utility.
    }
\end{table*}

\section{Value Bias and Lost Utility }
Dimensions of values and different types of utility, subjectiveness toward values and bounded rationality create gaps in perception of exchange and creation of values. Gaps create risks errors and opportunities, which depend on the observer and the subject of the gap. A good ethical behavior that expected from individuals from the perspective of the society is bridging gaps. Bridging gaps and avoid arbitrage by individuals creates value-driven progress and to avoid arbitrage.
Bad behavior for the society is exploitation of gaps by individuals to make a self-concentrated arbitrage. Arbitrage is an easy profit of value with no costs or risks from the perspective of an individual. However, the cost in arbitrage is always paid, in a different dimension. Perception and awareness of the society to the risks of gaps is critical for its survival and value preserving. Cost-optimal behavior with not full value aware judgment function is by default gaps/traps oriented because free is always better when optimizing cost. When we state that we are action cost-optimal we should continue and state on the expanse of what. In the next section, we discuss several aspects to address that question.

\subsection{The Forgotten Aspects of Problem Solving}
Most of the approach in the literature for planning, and in particular the goal-oriented approaches address only partially the real-world complexity. In the real-world scenarios, almost every plan has additional phases. When an agent that achieves his goals, sometimes he can stay at the location for which he planned, but most of the times he should retreat. Consider the well-known example of the logistic domain problem, which is addressed in many fields. The problem defined with packages to be delivered to various locations by various transportation methods as tracks and airplane. The goal in such domain usually defined by finding the most efficient (cost optimal) trajectories to match and deliver packages to pre-defined locations. Under the assumption of this definition, the problem is solved while the driver of the truck and the pilot at the location of the target. The retreat phase of planning is not addressed, which leaves the final phase of the problem in the hands of the people that defined as predicates implicitly as part of the truck or plane. If such plan reaches the industry, it brings damage to many values, human and financial, due to a limited perception of reality which compiled into academic advice for planning.

Suppose the goals are of a high priority and must be reached at any cost, we must stick to hard goals. A responsible solution for such scenarios should also plan the goal state and choose a plan to a goal within a state that will be suitable to retreat or to perform the following activity. If a goal can be reached from different trajectories, it will be responsible not to assume that the appearance of an agent in a state is equal regarding risk and values.

Solving complex systems with a diversity of values goals and restrictions, assuming a binary judgment function of achievements in a domain-independent approach consistent a paradox with many impacts on the user of the suggested solution. The efficiency that measured with such definition of achievements is an illusion of a success that obtained by simplifying complexity. Development of methods that are not applicable to real-world problems can create academic value, but we must be aware of the fact that in that case the utility is lost.

\section{Value Alignment}
\subsection{Retrospective}
Most of the papers report only final results, a ``snapshot of the finish line'', In total terms of efficiency of time and search for solution effort and the number of solved problems. Empirical evaluation based on a ``snapshot of the finish line" missing the entire story of the evaluated problem-solving method. When we measure complex and time-consuming solutions we can get much more information from the process of problem-solving, the behavior of the algorithm in different situations of complexity. Understanding why one problem solved while the other not on the process level rather than just reporting that. Each experiment can provide a retrospective on the proposed approach. Attention to measurement engineering of our research will lead to a better understanding and progress in research.

\subsection{ Relative Estimation of Achievements}
This paper suggests rethinking the usage of binary judgment functions of achievements for the sake of safety and effectiveness. Real-world problems are dynamic, and preferences depend on circumstances, while a perception of achievements as hard goals implies absolute preferences of values. Relative estimation methods are more flexible and realistic, particularly when a domain-independent approach for problem-solving discussed. The relativity of achievements and resource allowance must be considered more closely to scale up to real-world problem solving and domain-independent approaches.

Problems with limited resources and with a diversity of achievements are problems of choosing the most desirable within what available, choose within a budget and from what is available to choose, being aware to the fact that not all allowed. Achievements in such problems are relative by definition. Relative estimation approaches are the bridge between sparsity of problem-solving approaches and research fields to join forces to reduce risk and increase value. Net-benefit planning ~\cite{van2004effective,nigenda2005planning,baier2009heuristic,bonet:geffner:aij08,bonet1997robust1,colescoles:icaps11,keyder2009soft} is a relative estimation concerning achievements aware to costs in the process to achieve.

The nature of landmarks (which discussed in the following sections) allow to work in phases like cost preserving, investment, damage reducing value preserving; it is the natural bridge for synergistic combination and cooperation of approaches and researches toward a mutual target.

\section{Globally-Local Perspective}
Landmarks and milestones which are facts regarding the world or things the should be done along the way to a goal. The bridge for goal- and process-oriented approach is the reason for the recent success of the action landmarks. Based on their logical basis, the guidance obtained with action landmark implementation can be explained as breaking the search-process to local-goal-directed sub-processes. Local, as opposed to global optimization, is much more suitable for real-world problems complexity. What makes the landmarks to a bridge for goal- and process-oriented approaches is that the logical basis of action landmarks discovery is an end-process (goal) oriented. This observation allowed us to define the {\em end-process} oriented approach and leads us to the following conclusions;
\begin{enumerate}
 \item A general domain-independent planning should be defined concerning end-process oriented approach (based on the decision and rational decision-theoretic planning)
 \item The key for global optimization is in the optimization of landmarks discovery mechanism and exploitation with an objective to refine rationality regarding goal reachability.
\end{enumerate}
\subsubsection{A Rational Glimpse into the Future.}
Analysis of the problem allows deducing what must happen in each plan. Landmarks are the only rational information we have regarding the process of the problem. Extraction of a refined set of landmarks is a glimpse into the future, the allocation of the budget should be made upon that information. We should avoid unnecessary actions in case the budget is not sufficient to avoid waste of resources and effect on the environment. The first question that should be asked is ``can we do it effectively?", the answer should be rechecked along the plan.

{\bf Fact landmarks} are propositions that must be true at some point in every solution plan for a given planning task~\cite{HoffmannPS04}. {\em $\state$-landmark} is an assignment to a variable that is true at some point in {\em every} plan for a state $\state$.

{\bf Action landmarks} is an action $\action$ taken along every plan. Disjunctive action landmarks, each corresponding to a set of operators such that every plan contains at least one operator from that set~\cite{karpas.ijcai09,helmert:domshlak:icaps09,bonetH:ecai10,pommerening:13}.

{\bf Value driven landmarks} are an interrelated set of actions which confined to properties of plans that improve over the utility of a state $\state$ in hand~\cite{muller:Karpas:icaps18}. Treating achievements as landmarks and evaluation of the achievements in the context of the value that they represent, allows for a so-called improving approach for decision making in complex systems with over-subscribed goals or information. Empirical evaluations showed that planning and acting by with a target of self-improvement outperforms state-of-the-art goal-oriented approaches, namely collecting approaches. At a high level, an optimal plan consists of two parts: a prefix which achieves some valuable fact, and a suffix which maintains that valuable achievement, improving the utility of the other variables. The prefix can be thought of as ``investing" or being ``effective'', while the suffix can be thought of as ``reaping the benefits". Since these landmarks built to improve relatively instead achieve some goal or value absolutely, they are independent of the polarity of the utility value and apply to scenarios of decreasing damage as well as gaining a benefit. The relative estimation concept of value-driven-landmarks as an alternative perception of goals allows to address numerical and negative utilities.

{\bf Cost derived landmarks.} We define here a new set of landmarks that hold the information of time or location that an agent has to pay for external service. For example, pay for fuel or toll road. This information is derived from resources and constraints and not handled yet in the context of landmarks for reachability. Cost derived landmarks deduced from the cost of value-driven landmarks. First and necessary condition for an activity is that it leads to a substitutable, or improved value (effectively), which is represented by value-driven landmarks. Next, for an effective process only, cost derived (from effective value) landmark represent what and how many resources required to make it happen (efficiently). By analyzing budget and resource in a task that requires planning, a planner can recognize actions of pure lose if they are critical to keeping a process alive.

A process of refining and extraction of lengthier landmarks promise a better performance of AI agents. Landmarks are the only thing that is rational when an agent plan his actions.
\subsubsection{Rational Decision-Theoretic Planning.}
Traditional approaches for deterministic automated action planning, at its base, assumes rational-deterministic decision making without defining the bounds of rationality/determinism. We argue that the term of determinism has to be {\bf resource-bounded}, where time and processing capacity are the most basic bounds. Goals can be set, and planning approach should be chosen in a resource-bounded manner. Assuming unbounded-resource in hard problems is the first irrational decision and starts an irrational process of planning. That is the main reason that without changing the logical basis we will never scale-up to the real-world problems solving. It will be natural to combine the traditional deterministic approach with the decision-theoretic approaches for planning with uncertainty.

\textbf{Landmarks in Decision-Theoretic Planning.} Many decision-theoretic planning frameworks are focused on local reasoning and do not use landmarks heuristics. We suggest a dynamic goal as a reference which allows for landmark-based reasoning. This approach will improve local guidance based on information gained during the search and improve the performance of decision-theoretic solvers. Conceptually, these landmarks are the global guidance of local improvement process.

\subsubsection{The Value of Knowledge.}
Consider the value of knowledge; it is not clear how to evaluate knowledge. The nature of knowledge is such that can be evaluated in time perspective. Creating useful knowledge paid with the cost of all listed components and its value is shared and becomes in possession of the society, allows for evolution and wealth and progress. Many of the achievements in science and the history of ideas got their value only after the creator of the idea passed away. The other side of unfulfilled value potential of knowledge is involved crimes of stealing knowledge, plagiarism.
Costs paid in one dimension and value created in a different dimension, creating. A singularity of discovered observation appears with no costs behind. At the same time at the dimension of the origin of knowledge, there is an anomaly, expressed with multi-valued investment with no results as expected. Plagiarism is a real crime that has to be measured in a perspective of several dimensions. The Information Revolution and the increased speed of information transfer that came along with the development of the Internet this problem becomes acute and requires attention. If individuals in a society cannot trust society when they make long-term investments like research, long-term investments oriented with social values will be replaced by, short-term, cost-optimal investments. Another case of the value of knowledge is the entrepreneurship process where a group of individuals take a risk and make the long-term investment to give life to an idea. Even in a trusted system when the idea revealed to the market, there is a well-known phenomenon described as the time to market.
\subsubsection{Investment in knowledge.}
Investigation with an aim to create a useful and effective knowledge requires investment and involve risk, the risk of failure. Such investments should be encouraged. An individual in a trusted society can take the risk of long-term research if the society supports him and hedge the risk taken by the individual. Similarly to economic investment, a natural hedge of risk for research can be built on values such as the value of effective knowledge the value of failure. If society can make value out of failure, it will lead to a more significant benefit concerning academy goals along with additional benefits on several dimensions. Hedge for a long-term investment in sustainable values, human values. Long-term investments are proved to be an excellent strategy to create an effective value. Applying, long-term investment strategy, we suggest combining observations from AI problem solving, planning and search in large-scale domain approaches. Search in large-scale domains and research are very related to each other not only by name but also in the concept and dynamics of making a quick guess on a path within and to uncertainty. Inspired by these similarities and we suggest another hypothesis to estimate values. The relation of trust in the community to publishing failures as a key to collaborative and effective research progress. The branch-and-bound pruning algorithm considered being effective in complex over-subscription planning systems~\cite{bonet1997robust1,bonet1997robust2,mirkis:domshlak:jair15,muller:Karpas:icaps18}, is in practice make value of failing by using mistakes to speed up the progress with the pruning method is applied for branches that cannot promise a better node in the branch of interest. Since search and research are similar in many ways, there is a place to investigate this branch more closely to exploit also what we know about what is wrong.
\section{Discussion}
In these days, with the rapid advance of AI decision-making systems, the researcher and the developer are the gates to make or stop the crimes in values. It is a more important duty of the developer to analyze the risks of his algorithm than coding or writing grammatically perfect code, and it is even more important to understand the logic and explain the user the dangerous of the proposed AI solutions and systems. Recent studies on {\em explainable AI}~\cite{DBLP:journals/corr/abs-1709-10256,gunning2017explainable,borgo2018towards,vigan2018explainable,doran2017does,miller2017explanation} recognized the gap and made an important step to bridge it by taking responsibility for explaining AI decision nature. \cite{mittelstadt2018explaining} Explaining and collection of feedback from the environment is an ongoing and infinite process that should be in the ethics' basis of a developer and researcher. It is a responsibility that should be integrated into any process of creation. When we make assumptions of perfect or unharmful outcomes in the development process, it is by default comes at the expense of the user. The importance of explaining to AI professionals recently raised~\cite{mittelstadt2018explaining}, showing the impact and risks of encoded decisions on end-user. Being unfamiliar with the user makes it easier to cut corner since it is an easy benefit of time to the developer which even cannot imagine the subject of his crimes. It will be natural to start the regulation from the system and lead a change of principles, to ask a paper to state its utilities and values, which values are improved and which might be in danger.

It is somehow accepted that a process of writing a paper is done with its publication, and what published is unpublishable. At the same time, there is no feedback aggregated over time. A retrospective and responsible maintenance of ideas must be encouraged. The user must have a place to update feedback from the field. Without maintaining feedback, the academic system will lose the trust of society. Without the interest of the society, the research community will solve non-real-world problems, blinded by many publications, fail to see that it is a bobble --- not a real-world problem.

\bibliographystyle{aaai}
\bibliography{muller}

\begin{thebibliography}{}

\bibitem[\protect\citeauthoryear{Ackoff}{1989}]{ackoff1989data}
Ackoff, R.~L.
\newblock 1989.
\newblock From data to wisdom.
\newblock {\em Journal of applied systems analysis} 16(1):3--9.

\bibitem[\protect\citeauthoryear{Aghighi and
  Jonsson}{2014}]{aghighi2014oversubscription}
Aghighi, M., and Jonsson, P.
\newblock 2014.
\newblock Oversubscription planning: Complexity and compilability.
\newblock In {\em AAAI},  2221--2227.

\bibitem[\protect\citeauthoryear{Baier, Bacchus, and
  McIlraith}{2009}]{baier2009heuristic}
Baier, J.~A.; Bacchus, F.; and McIlraith, S.~A.
\newblock 2009.
\newblock A heuristic search approach to planning with temporally extended
  preferences.
\newblock {\em Artificial Intelligence} 173(5):593.

\bibitem[\protect\citeauthoryear{Benton, Do, and
  Kambhampati}{2009}]{bonet1997robust1}
Benton, J.; Do, M.; and Kambhampati, S.
\newblock 2009.
\newblock Anytime heuristic search for partial satisfaction planning.
\newblock {\em Artificial Intelligence} 173(5-6):562--592.

\bibitem[\protect\citeauthoryear{Benton}{2006}]{benton2006solving}
Benton, J.
\newblock 2006.
\newblock Solving goal utility dependencies and simple preferences in partial
  satisfaction planning.
\newblock {\em ICAPS 2006} ~16.

\bibitem[\protect\citeauthoryear{Berger}{2013}]{berger2013statistical}
Berger, J.~O.
\newblock 2013.
\newblock {\em Statistical decision theory and Bayesian analysis}.
\newblock Springer Science \& Business Media.

\bibitem[\protect\citeauthoryear{Bonet and Geffner}{2008}]{bonet:geffner:aij08}
Bonet, B., and Geffner, H.
\newblock 2008.
\newblock Heuristics for planning with penalties and rewards formulated in
  logic and computed through circuits.
\newblock {\em Artificial Intelligence} 172(12-13):1579--1604.

\bibitem[\protect\citeauthoryear{Bonet and Helmert}{2010}]{bonetH:ecai10}
Bonet, B., and Helmert, M.
\newblock 2010.
\newblock Strengthening landmark heuristics via hitting sets.
\newblock In {\em ECAI},  329--334.

\bibitem[\protect\citeauthoryear{Bonet, Loerincs, and
  Geffner}{1997}]{bonet1997robust2}
Bonet, B.; Loerincs, G.; and Geffner, H.
\newblock 1997.
\newblock A robust and fast action selection mechanism for planning.
\newblock In {\em AAAI/IAAI},  714--719.

\bibitem[\protect\citeauthoryear{Borgo, Cashmore, and
  Magazzeni}{2018}]{borgo2018towards}
Borgo, R.; Cashmore, M.; and Magazzeni, D.
\newblock 2018.
\newblock Towards providing explanations for ai planner decisions.
\newblock {\em arXiv preprint arXiv:1810.06338}.

\bibitem[\protect\citeauthoryear{Boutilier, Dean, and
  Hanks}{1999}]{boutilier1999decision}
Boutilier, C.; Dean, T.; and Hanks, S.
\newblock 1999.
\newblock Decision-theoretic planning: Structural assumptions and computational
  leverage.
\newblock {\em Journal of Artificial Intelligence Research} 11:1--94.

\bibitem[\protect\citeauthoryear{Coles and Coles}{2011}]{colescoles:icaps11}
Coles, A.~J., and Coles, A.
\newblock 2011.
\newblock Lprpg-p: Relaxed plan heuristics for planning with preferences.
\newblock In {\em ICAPS}.

\bibitem[\protect\citeauthoryear{Cooley}{1980}]{cooley1980architect}
Cooley, M.
\newblock 1980.
\newblock {\em Architect or bee?}
\newblock Langley Technical Services Slough.

\bibitem[\protect\citeauthoryear{Cunliffe}{2008}]{cunliffe2008orientations}
Cunliffe, A.~L.
\newblock 2008.
\newblock Orientations to social constructionism: Relationally responsive
  social constructionism and its implications for knowledge and learning.
\newblock {\em Management Learning} 39(2):123--139.

\bibitem[\protect\citeauthoryear{DeGroot}{2005}]{degroot2005optimal}
DeGroot, M.~H.
\newblock 2005.
\newblock {\em Optimal statistical decisions}, volume~82.
\newblock John Wiley \& Sons.

\bibitem[\protect\citeauthoryear{Do and Kambhampati}{2004}]{do2004partial}
Do, M., and Kambhampati, S.
\newblock 2004.
\newblock Partial satisfaction (over-subscription) planning as heuristic
  search.
\newblock {\em Proceedings of KBCS-04}.

\bibitem[\protect\citeauthoryear{Do \bgroup et al\mbox.\egroup
  }{2007}]{DoBBK:ijcai07}
Do, M.~B.; Benton, J.; Van Den~Briel, M.; and Kambhampati, S.
\newblock 2007.
\newblock Planning with goal utility dependencies.
\newblock In {\em IJCAI},  1872--1878.

\bibitem[\protect\citeauthoryear{Domshlak and
  Mirkis}{2015}]{mirkis:domshlak:jair15}
Domshlak, C., and Mirkis, V.
\newblock 2015.
\newblock Deterministic oversubscription planning as heuristic search:
  Abstractions and reformulations.
\newblock {\em Journal of Artificial Intelligence Research} 52:97--169.

\bibitem[\protect\citeauthoryear{Doran, Schulz, and
  Besold}{2017}]{doran2017does}
Doran, D.; Schulz, S.; and Besold, T.~R.
\newblock 2017.
\newblock What does explainable ai really mean? a new conceptualization of
  perspectives.
\newblock {\em arXiv preprint arXiv:1710.00794}.

\bibitem[\protect\citeauthoryear{Eliot}{1934}]{eliot1934rock}
Eliot, T.~S.
\newblock 1934.
\newblock {\em The Rock: A Pageant Play, Written for Performance at Sadler's
  Wells Theatre, 28 May-9 June 1934, on Behalf of the Forty-five Churches Fund
  of the Dioceses of London; Book of Words by TS Eliot}.
\newblock Faber \& Faber.

\bibitem[\protect\citeauthoryear{Empson}{2001}]{empson2001fear}
Empson, L.
\newblock 2001.
\newblock Fear of exploitation and fear of contamination: Impediments to
  knowledge transfer in mergers between professional service firms.
\newblock {\em Human relations} 54(7):839--862.

\bibitem[\protect\citeauthoryear{Farber, Costanza, and
  Wilson}{2002}]{farber2002economic}
Farber, S.~C.; Costanza, R.; and Wilson, M.~A.
\newblock 2002.
\newblock Economic and ecological concepts for valuing ecosystem services.
\newblock {\em Ecological economics} 41(3):375--392.

\bibitem[\protect\citeauthoryear{Fox, Long, and
  Magazzeni}{2017}]{DBLP:journals/corr/abs-1709-10256}
Fox, M.; Long, D.; and Magazzeni, D.
\newblock 2017.
\newblock Explainable planning.
\newblock {\em CoRR} abs/1709.10256.

\bibitem[\protect\citeauthoryear{Garrison, Mestre-Ferrandiz, and
  Zamora}{2016}]{garrison2016value}
Garrison, L.; Mestre-Ferrandiz, J.; and Zamora, B.
\newblock 2016.
\newblock The value of knowing and knowing the value: improving the health
  technology assessment of complementary diagnostics.
\newblock {\em Office of Health Economics: London, UK}.

\bibitem[\protect\citeauthoryear{Gr{\"u}ne-Yanoff}{2012}]{grune2012paradoxes}
Gr{\"u}ne-Yanoff, T.
\newblock 2012.
\newblock Paradoxes of rational choice theory.
\newblock In {\em Handbook of Risk Theory}. Springer.
\newblock  499--516.

\bibitem[\protect\citeauthoryear{Gunning}{2017}]{gunning2017explainable}
Gunning, D.
\newblock 2017.
\newblock Explainable artificial intelligence (xai).
\newblock {\em Defense Advanced Research Projects Agency (DARPA), nd Web}.

\bibitem[\protect\citeauthoryear{Helmert and
  Domshlak}{2009}]{helmert:domshlak:icaps09}
Helmert, M., and Domshlak, C.
\newblock 2009.
\newblock Landmarks, critical paths and abstractions: what's the difference
  anyway?
\newblock In {\em ICAPS},  162--169.

\bibitem[\protect\citeauthoryear{Hoffmann, Porteous, and
  Sebastia}{2004}]{HoffmannPS04}
Hoffmann, J.; Porteous, J.; and Sebastia, L.
\newblock 2004.
\newblock Ordered landmarks in planning.
\newblock {\em Journal of Artificial Intelligence Research} 22:215--278.

\bibitem[\protect\citeauthoryear{Karpas and Domshlak}{2009}]{karpas.ijcai09}
Karpas, E., and Domshlak, C.
\newblock 2009.
\newblock Cost-optimal planning with landmarks.
\newblock In {\em IJCAI},  1728--1733.

\bibitem[\protect\citeauthoryear{Keyder and Geffner}{2009}]{keyder2009soft}
Keyder, E., and Geffner, H.
\newblock 2009.
\newblock Soft goals can be compiled away.
\newblock {\em Journal of Artificial Intelligence Research} 36:547--556.

\bibitem[\protect\citeauthoryear{Kriger and
  Barnes}{1992}]{kriger1992organizational}
Kriger, M.~P., and Barnes, L.~B.
\newblock 1992.
\newblock Organizational decision-making as hierarchical levels of drama.
\newblock {\em Journal of Management Studies} 29(4):439--457.

\bibitem[\protect\citeauthoryear{Marwala}{2013}]{marwala2013flexibly}
Marwala, T.
\newblock 2013.
\newblock Flexibly-bounded rationality and marginalization of irrationality
  theories for decision making.
\newblock {\em arXiv preprint arXiv:1306.2025}.

\bibitem[\protect\citeauthoryear{Marwala}{2014}]{marwala2014artificial}
Marwala, T.
\newblock 2014.
\newblock {\em Artificial intelligence techniques for rational decision
  making}.
\newblock Springer.

\bibitem[\protect\citeauthoryear{Marwala}{2015}]{marwala2015causality}
Marwala, T.
\newblock 2015.
\newblock {\em Causality, correlation and artificial intelligence for rational
  decision making}.
\newblock World Scientific.

\bibitem[\protect\citeauthoryear{Miller}{2017}]{miller2017explanation}
Miller, T.
\newblock 2017.
\newblock Explanation in artificial intelligence: insights from the social
  sciences.
\newblock {\em arXiv preprint arXiv:1706.07269}.

\bibitem[\protect\citeauthoryear{Mittelstadt, Russell, and
  Wachter}{2018}]{mittelstadt2018explaining}
Mittelstadt, B.; Russell, C.; and Wachter, S.
\newblock 2018.
\newblock Explaining explanations in ai.
\newblock {\em arXiv preprint arXiv:1811.01439}.

\bibitem[\protect\citeauthoryear{Muller and
  Karpas}{2018}]{muller:Karpas:icaps18}
Muller, D., and Karpas, E.
\newblock 2018.
\newblock Value driven landmarks for oversubscription planning.
\newblock In {\em ICAPS}.

\bibitem[\protect\citeauthoryear{Nigenda and
  Kambhampati}{2005}]{nigenda2005planning}
Nigenda, R.~S., and Kambhampati, S.
\newblock 2005.
\newblock Planning graph heuristics for selecting objectives in
  over-subscription planning problems.
\newblock In {\em ICAPS},  192--201.

\bibitem[\protect\citeauthoryear{Poladian \bgroup et al\mbox.\egroup
  }{2003}]{poladian2003time}
Poladian, V.; Butler, S.; Shaw, M.; and Garlan, D.
\newblock 2003.
\newblock Time is not money: The case for multi-dimensional accounting in
  value-based software engineering.
\newblock In {\em Proceedings of the 5th Int’l Workshop on Economics Driven
  Software Engineering Research (EDSER-5}.
\newblock Citeseer.

\bibitem[\protect\citeauthoryear{Pommerening and
  Helmert}{2013}]{pommerening:13}
Pommerening, F., and Helmert, M.
\newblock 2013.
\newblock Incremental lm-cut.
\newblock In {\em ICAPS}.

\bibitem[\protect\citeauthoryear{Rowley}{2007}]{Jennifer2007Rowley}
Rowley, J.
\newblock 2007.
\newblock The wisdom hierarchy: representations of the dikw hierarchy.
\newblock {\em Journal of Information Science} 33(2):163--180.

\bibitem[\protect\citeauthoryear{Simon}{1957}]{simon1957models}
Simon, H.~A.
\newblock 1957.
\newblock Models of man; social and rational.

\bibitem[\protect\citeauthoryear{Simon}{1990}]{simon1990mechanism}
Simon, H.~A.
\newblock 1990.
\newblock A mechanism for social selection and successful altruism.
\newblock {\em Science(Washington)} 250(4988):1665--1668.

\bibitem[\protect\citeauthoryear{Simon}{1991}]{simon1991bounded}
Simon, H.~A.
\newblock 1991.
\newblock Bounded rationality and organizational learning.
\newblock {\em Organization science} 2(1):125--134.

\bibitem[\protect\citeauthoryear{Smith}{1776}]{smith1776inquiry}
Smith, A.
\newblock 1776.
\newblock An inquiry into the nature and causes of the wealth of nations:
  Volume one.
\newblock London: printed for W. Strahan; and T. Cadell, 1776.

\bibitem[\protect\citeauthoryear{Smith}{2004}]{smith:icaps04}
Smith, D.~E.
\newblock 2004.
\newblock Choosing objectives in over-subscription planning.
\newblock In {\em ICAPS}, volume~4,  393.

\bibitem[\protect\citeauthoryear{Spender}{1996}]{spender1996organizational}
Spender, J.-C.
\newblock 1996.
\newblock Organizational knowledge, learning and memory: three concepts in
  search of a theory.
\newblock {\em Journal of organizational change management} 9(1):63--78.

\bibitem[\protect\citeauthoryear{Strati}{2007}]{strati2007sensible}
Strati, A.
\newblock 2007.
\newblock Sensible knowledge and practice-based learning.
\newblock {\em Management learning} 38(1):61--77.

\bibitem[\protect\citeauthoryear{Swart}{2011}]{knowingCreatesValue2011}
Swart, J.
\newblock 2011.
\newblock That’s why it matters: How knowing creates value.
\newblock {\em Management Learning} 42(3):319--332.

\bibitem[\protect\citeauthoryear{Van Den~Briel \bgroup et al\mbox.\egroup
  }{2004}]{van2004effective}
Van Den~Briel, M.; Sanchez, R.; Do, M.~B.; and Kambhampati, S.
\newblock 2004.
\newblock Effective approaches for partial satisfaction (over-subscription)
  planning.
\newblock In {\em AAAI},  562--569.

\bibitem[\protect\citeauthoryear{Van Den~Briel, Sanchez, and
  Kambhampati}{2004}]{van2004over}
Van Den~Briel, M.; Sanchez, R.; and Kambhampati, S.
\newblock 2004.
\newblock Over-subscription in planning: A partial satisfaction problem.
\newblock In {\em ICAPS-04 Workshop on Integrating Planning into Scheduling},
  91--98.

\bibitem[\protect\citeauthoryear{Viganò and
  Magazzeni}{2018}]{vigan2018explainable}
Viganò, L., and Magazzeni, D.
\newblock 2018.
\newblock Explainable security.

\bibitem[\protect\citeauthoryear{Zeleny}{1987}]{zeleny1987management}
Zeleny, M.
\newblock 1987.
\newblock Management support systems: Towards integrated knowledge management.
\newblock {\em Human systems management} 7(1):59--70.

\end{thebibliography}
\end{document}